\documentclass[manuscript,nonacm]{acmart}

\AtBeginDocument{%
  \providecommand\BibTeX{{%
    \normalfont B\kern-0.5em{\scshape i\kern-0.25em b}\kern-0.8em\TeX}}}

\begin{document}

\title{Sophisticated Students in Boston Mechanism and Gale-Shapley Algorithm for School Choice Problem}

\author{Abhishek Paudel}
\affiliation{%
  \institution{George Mason University}
  \city{Fairfax}
  \state{Virginia}
  \country{USA}
}


\begin{abstract}
  
We present our experimental results of simulating the school choice problem which deals with the assignment of students to schools based on each group's complete preference list for the other group using two algorithms: Boston mechanism and student-proposing Gale-Shapley algorithm. We compare the effects of sophisticated students altering their preference lists with regards to these two algorithms. Our simulation results show that sophisticated students can benefit more in Boston mechanism compared to Gale-Shapley algorithm based on multiple evaluation metrics.
\end{abstract}



\maketitle
\pagestyle{plain} 

\section{Introduction}

Finding a set of matches between students and schools is an interesting problem commonly known as the \textit{school choice problem}. Various algorithms have been used to find best matching between students and schools in real world scenario. In this paper, we study two variants of algorithms that are used to solve this problem: Boston mechanism also known as \textit{immediate acceptance algorithm} and student-proposing Gale-Shapley algorithm also known as \textit{deferred acceptance algorithm}. We introduce the school choice problem as follows.

\subsection{School Choice Problem}
The school choice problem\cite{abdul2003} consists of the following.

\begin{itemize}
    \item A set of $n$ students $A = \{a_1, a_2,...a_n \}$.
    \item A set of $m$ schools $B = \{b_1, b_2,...b_m \}$.
    \item Each school $b_j \in B$ has a capacity of $q_j$ which is the maximum number of students it can admit.
    \item Each student $a_i \in A$ has a strict preference list $P_i$ that ranks all schools in $B$. The set of all students' preference lists is $P= \{P_1, P_2,...P_n\}$
    \item Each school $b_j \in B$ has a strict preference list $Q_j$ that ranks all students in $A$. The set of all schools' preference lists is $Q= \{Q_1, Q_2,...Q_m\}$
\end{itemize}

The task is to find a matching $M: A \rightarrow B$ such that each student $a_i$ is matched to one and only one school, and each school $b_j$ is matched to no more than $q_j$ students. By taking into account the preference lists of schools and students, different algorithms can find different sets of matches with different characteristics.

\subsection{Boston Mechanism}
Boston mechanism is one of the widely used matching algorithms used in the United States\cite{miralles2009school}. The Boston mechanism tries to match as many students to their most preferred school as possible. In each round $k$, we take all unassigned students and take the schools with $k$th priority in these students' priority lists. Then, these schools are assigned to the students based on the school's priority over the students. If the school hasn't reached its maximum capacity, the student is admitted, else the student remains unassigned in this round and will be attempted to match again in the next round. This is repeated until no student remains unassigned. Note that the maximum number of rounds we need to iterate over until no students remain unassigned is equal to the number of schools assuming that the cumulative capacity of all schools is greater than or equal to the total number of students.

\subsection{Gale-Shapley Algorithm}
Gale-Shapley algorithm\cite{gale1962} is known for its stable matching properties which is desired in many of such matching problems. In the student-proposing variant of the Gale-Shapley algorithm, each student picks their most preferred school that hasn't already rejected them previously. If the school hasn't already reached its capacity, the student is provisionally admitted to the school. However, if the school has reached its capacity, the school looks at its priority list over the students. If this student has higher priority than any of the provisionally admitted students, then the admitted student with lowest priority is unassigned from the school, the new student with higher priority is admitted in its place. Once there remains no unassigned students, the obtained provisional assignments become the final assignments. Note that we assume that the cumulative capacity of all schools is greater than or equal to the total number of students.

\subsection{Sophisticated and Sincere Students}
The two algorithms described above result in different sets of matches between students and schools. And depending upon the preference lists of students and schools, the matches generated by each of these algorithms can also be different. This suggests that by altering the preference lists over schools, a student can get admitted to a different school than they would otherwise be admitted to if they hadn't altered preferences\cite{pathak2008}. We call such students sophisticated students who will alter their preferences so as to get admitted to a different school which might otherwise have not been possible if they didn't alter their preference. Sincere students are those who will report their true preferences.

\section{Methodology}
We aim to study the effects of sophisticated students altering their preference so as to benefit in the process by experimentally simulating the school choice problem. We generate a set of students, a set of schools with certain capacities, and their preferences. Then we use Boston mechanism and student-proposing Gale-Shapley algorithm to find the matches. Next, we randomly sample certain fraction of the students as sophisticated students who will alter their preference lists using various strategies. We then obtain new sets of matches corresponding to these new preferences using Boston mechanism and student-proposing Gale-Shapley algorithm. Finally, we evaluate the properties of resulting matches obtained after preference alteration and compare them between both algorithms.

\subsection{Preference List Generation}
We generate preference lists for students and schools by taking into consideration various factors. The preference lists of students for schools are generated based on the following four features: distance to schools, whether they have a sibling studying at a school, school tier and random factor.
The preference lists of schools are generated based only upon the distance to students. For schools, we discretize the generated preference values based on distance into four values 1, 2, 3 and 4 where lower values denotes higher preference of a student in a school. For generating the distances, we first randomly generate students and schools in a 2D plane where \textit{x} coordinates and \textit{y} coordinates are uniformly sampled between 0 and 1. Then, we calculate the pairwise distance between them. Next, we sample whether a student has a sibling or not based on Bernoulli distribution, and assign this sibling to random school. The school tiers are sampled with varying probabilities from tier-1, tier-2, tier-3 or tier-4 where tier-1 represents a top tier school and tier-4 represents a bottom tier school. Random factor is uniformly sampled between 0 and 1.

After generating the four features as described above, we combine all these features to generate the preference values of students for each school. The preference value is generated by normalizing the above features appropriately and taking weighted sum of these features. A lower preference value for a school means that the student prefers that school more compared to another school with a higher preference value.

\subsection{Preference List Alteration}
We sample a varying fraction of sophisticated and sincere students from the student pool. We choose sophisticated students by randomly choosing students from the student pool without replacement until desired number of sophisticated students are sampled. Sophisticated students will then have their preference lists altered using various strategies. We employ three strategies that sophisticated students use to alter their preferences.

\begin{enumerate}
    \item \textit{Strategy A}: Find a school that most students have on rank 1. And if you also have that school on rank 1, swap schools on rank 1 and rank 2.
    \item \textit{Strategy B}: Find a school that most students have on rank 1. And if you also have that school on rank 1, move that school to the last in your preference list.
    \item \textit{Strategy C}: Find a school that most students have on the bottom half of their preference lists (a proxy for less popular school). And if you have that school on top half of your preference list, move that school to rank 1. 
\end{enumerate}

\subsection{Evaluation Metrics}

We use the following evaluation metrics as appropriate based on the preference alteration strategies used.

\begin{enumerate}
    \item \textit{EM-Top3}: The percentage of sophisticated students that got admitted to a school that is on top 3 in their true preference list after they altered their preference list. Before altering preference, they would not be admitted to their top 3 schools.
    \item \textit{EM-Higher}: The percentage of sophisticated students that got admitted to a school that is ranked higher in their true preference list after they altered their preference list.
    \item \textit{EM-Selected}: The percentage of sophisticated students that got admitted to a specific selected school after they altered their preference list. This metric is specifically useful for Strategy C with the less popular school as selected school. 
\end{enumerate}

\section{Experiments}
We run our experiments for 100 simulations with the number of students fixed to 2000 and number of schools fixed to 20 for all our simulation instances with different instances of preference lists of students and schools and aggregate the results from all simulations. For each instance of simulation, school capacities are generated randomly for all schools such that cumulative capacities of all schools are equal to the total number of students. We then randomly generate preference lists for students and schools as described in the previous section. The preference values of schools are discretized into 1, 2, 3 and 4 corresponding to the bins 0, 0.3, 0.5, 0.7 and 1 respectively so that each student has a priority of 1, 2, 3 or 4 where 1 denotes highest priority and 4 denotes lowest priority. Each feature for preference list generation of students are weighted as follows: distance to schools (0.5), sibling (0.2), school tier (0.2) and random factor (0.1). The Bernoulli distribution for sampling siblings uses the success probability of 0.5 meaning that approximately half of the students have a sibling in one of the schools. For sampling the school tiers, we sample tier-1, tier-2, tier-3 and tier-4 schools with probabilities 0.1, 0.2, 0.3 and 0.4 respectively.

\subsection{Strategy A}
\begin{figure}
    \centering
    \includegraphics[width=11cm]{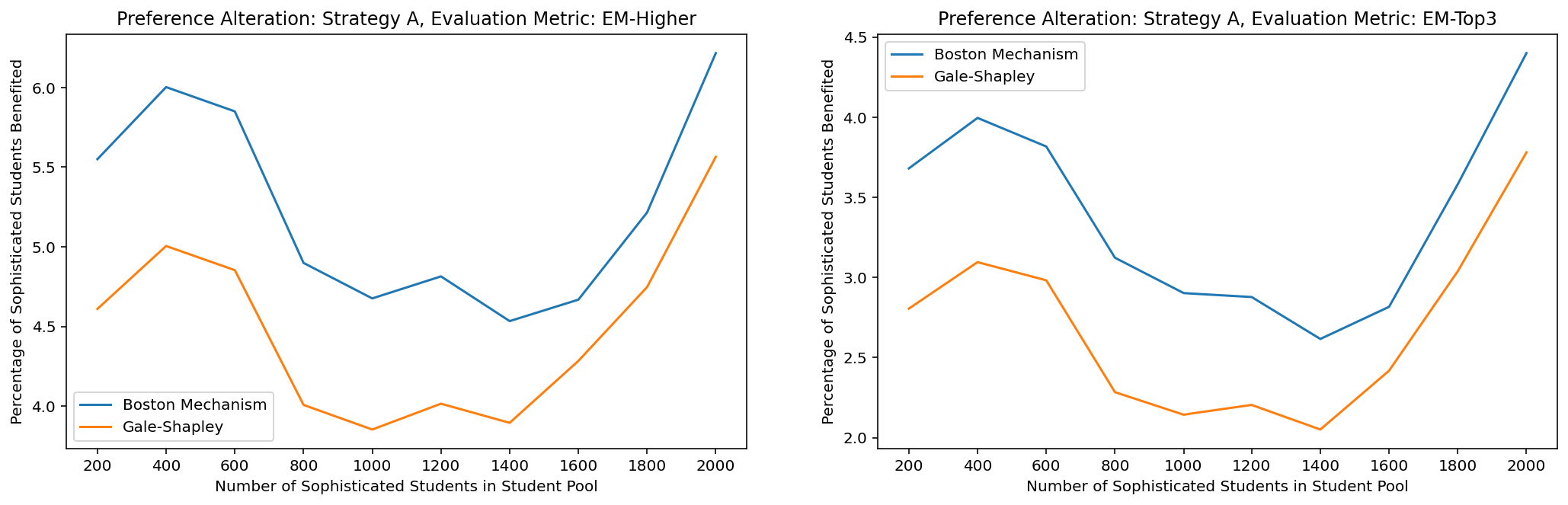}
    \caption{Percentage of sophisticated students who got admitted to a higher ranked school (left) and their top-3 school (right) after altering their preferences using Strategy A (Results are averaged over 100 trials)}
    \label{fig:strategy_a}
\end{figure}

The results for preference alteration using Strategy A are shown in Figure \ref{fig:strategy_a}. We observe similar trends for both evaluation metrics EM-Higher and EM-Top3 showing that more percentage of sophisticated students are able to benefit in Boston mechanism compared to Gale-Shapley algorithm. This remains true for any number of sophisticated students in the student pool. We also observe that as the number of sophisticated students increases, less percentage of them are able to benefit using this strategy. However, after the number of sophisticated students increases to more than 1400, more fraction of them are benefited. This could be because when a large number of students prefer one school over other, there are going to be some sophisticated students who will have preferred more a school that other students preferred less, and hence are likely to get admitted to that school.

\subsection{Strategy B}
\begin{figure}
    \centering
    \includegraphics[width=11cm]{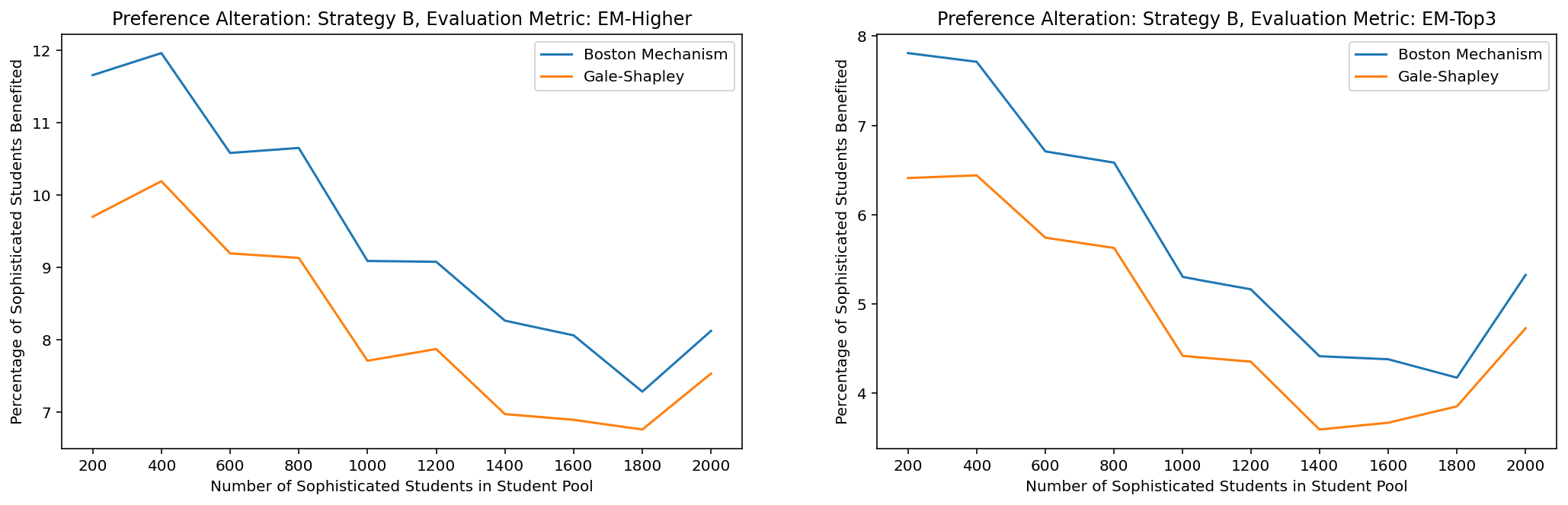}
    \caption{Percentage of sophisticated students who got admitted to a higher ranked school (left) and their top-3 school (right) after altering their preferences using Strategy B (Results are averaged over 100 trials)}
    \label{fig:strategy_b}
\end{figure}

Results for preference alteration using Strategy B are shown in Figure \ref{fig:strategy_b}. We observe that more percentage of sophisticated are benefited when there the number of sophisticated students is small. We also see that Boston mechanism is more beneficial to sophisticated students compared to Gale-Shapley algorithm. As the number of sophisticated students starts to increase in the student pool, less and less fraction of them are able to benefit until except all students become sophisticated at which point there is slight increase in benefited fraction of sophisticated students likely for the same reason as with Strategy A explained earlier.

\subsection{Strategy C}

\begin{figure}
    \centering
    \includegraphics[width=6cm]{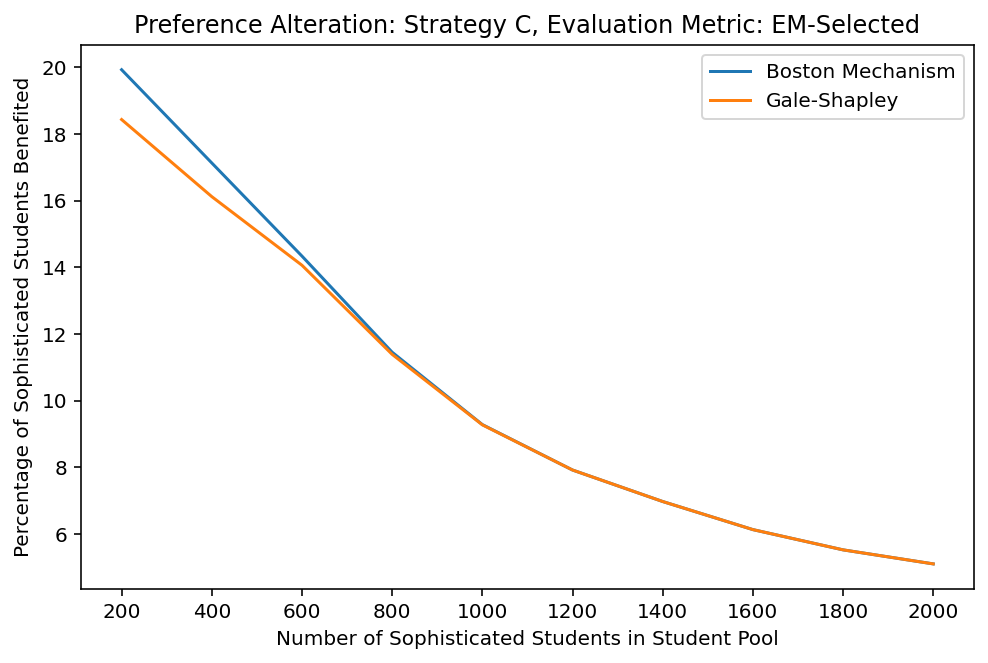}
    \caption{Percentage of sophisticated students who got admitted to a selected less popular school (EM-Selected) after altering their preferences using Strategy C (Results are averaged over 100 trials)}
    \label{fig:strategy_c_em_selected}
\end{figure}

This strategy is somewhat of a proxy for finding a less popular school that one likes and putting that school on rank 1. We evaluate this strategy based on what fraction of sophisticated students are able to get admitted to this less popular school that they like as represented by the evaluation metric EM-Selected with the selected school being the less popular school that they like. The results are shown in \ref{fig:strategy_c_em_selected}. With this strategy, we see a similar trend where as the number of sophisticated students increase, lesser fraction of them are benefited with this strategy. However, this strategy benefits the most compared to other two strategies as seen in \ref{fig:strategy_c_em_selected} which starts at around 18\% to 20\% sophisticated students being benefited when there are around 200 sophisticated students. Interestingly, the gap between Boston mechanism and Gale-Shapley algorithm closes immediately when number of sophisticated students starts to increase with both algorithms being equally likely to be beneficial for sophisticated students using Strategy C.

\section{Conclusion and Future Direction}
Overall, we see that Boston mechanism is more beneficial to sophisticated students than Gale-Shapley algorithm in almost all of our experiments. This suggests Gale-Shapley algorithm is more robust when it comes to sophisticated students trying to game the matching mechanism. We also find that in general, lesser the number of sophisticated students in the student pool, more likely they will benefit by altering their preferences. Additionally , this is also more true with Boston mechanism than with Gale-Shapley algorithm. The future direction of this research could be headed to study the effects of preference alteration by sophisticated students in sincere students. It would also be interesting study various other types of matching algorithms that are derived from Boston mechanism and Gale-Shapley algorithm for various matching settings in real world scenario.

\bibliographystyle{ACM-Reference-Format}
\bibliography{main}


\begin{thebibliography}{4}


\ifx \showCODEN    \undefined \def \showCODEN     #1{\unskip}     \fi
\ifx \showDOI      \undefined \def \showDOI       #1{#1}\fi
\ifx \showISBNx    \undefined \def \showISBNx     #1{\unskip}     \fi
\ifx \showISBNxiii \undefined \def \showISBNxiii  #1{\unskip}     \fi
\ifx \showISSN     \undefined \def \showISSN      #1{\unskip}     \fi
\ifx \showLCCN     \undefined \def \showLCCN      #1{\unskip}     \fi
\ifx \shownote     \undefined \def \shownote      #1{#1}          \fi
\ifx \showarticletitle \undefined \def \showarticletitle #1{#1}   \fi
\ifx \showURL      \undefined \def \showURL       {\relax}        \fi
\providecommand\bibfield[2]{#2}
\providecommand\bibinfo[2]{#2}
\providecommand\natexlab[1]{#1}
\providecommand\showeprint[2][]{arXiv:#2}

\bibitem[\protect\citeauthoryear{Abdulkadiroğlu and Sönmez}{Abdulkadiroğlu
  and Sönmez}{2003}]%
        {abdul2003}
\bibfield{author}{\bibinfo{person}{Atila Abdulkadiroğlu} {and}
  \bibinfo{person}{Tayfun Sönmez}.} \bibinfo{year}{2003}\natexlab{}.
\newblock \showarticletitle{School Choice: A Mechanism Design Approach}.
\newblock \bibinfo{journal}{\emph{American Economic Review}}
  \bibinfo{volume}{93}, \bibinfo{number}{3} (\bibinfo{date}{June}
  \bibinfo{year}{2003}), \bibinfo{pages}{729--747}.
\newblock
\urldef\tempurl%
\url{https://doi.org/10.1257/000282803322157061}
\showDOI{\tempurl}


\bibitem[\protect\citeauthoryear{Gale and Shapley}{Gale and Shapley}{1962}]%
        {gale1962}
\bibfield{author}{\bibinfo{person}{D. Gale} {and} \bibinfo{person}{L.~S.
  Shapley}.} \bibinfo{year}{1962}\natexlab{}.
\newblock \showarticletitle{College Admissions and the Stability of Marriage}.
\newblock \bibinfo{journal}{\emph{The American Mathematical Monthly}}
  \bibinfo{volume}{69}, \bibinfo{number}{1} (\bibinfo{date}{Jan.}
  \bibinfo{year}{1962}), \bibinfo{pages}{9}.
\newblock
\urldef\tempurl%
\url{https://doi.org/10.2307/2312726}
\showDOI{\tempurl}


\bibitem[\protect\citeauthoryear{Miralles}{Miralles}{2009}]%
        {miralles2009school}
\bibfield{author}{\bibinfo{person}{Antonio Miralles}.}
  \bibinfo{year}{2009}\natexlab{}.
\newblock \showarticletitle{School choice: The case for the Boston mechanism}.
  In \bibinfo{booktitle}{\emph{International conference on auctions, market
  mechanisms and their applications}}. Springer, \bibinfo{pages}{58--60}.
\newblock


\bibitem[\protect\citeauthoryear{Pathak and Sönmez}{Pathak and
  Sönmez}{2008}]%
        {pathak2008}
\bibfield{author}{\bibinfo{person}{Parag~A. Pathak} {and}
  \bibinfo{person}{Tayfun Sönmez}.} \bibinfo{year}{2008}\natexlab{}.
\newblock \showarticletitle{Leveling the Playing Field: Sincere and
  Sophisticated Players in the Boston Mechanism}.
\newblock \bibinfo{journal}{\emph{American Economic Review}}
  \bibinfo{volume}{98}, \bibinfo{number}{4} (\bibinfo{date}{September}
  \bibinfo{year}{2008}), \bibinfo{pages}{1636--52}.
\newblock
\urldef\tempurl%
\url{https://doi.org/10.1257/aer.98.4.1636}
\showDOI{\tempurl}


\end{thebibliography}


\end{document}